\documentclass[runningheads]{llncs}
\usepackage{comment}
\usepackage[T1]{fontenc}
\usepackage{booktabs}
\usepackage{amsmath}
\usepackage{amssymb}
\usepackage{graphicx}
\usepackage{multicol}
\usepackage{multirow}
\usepackage{adjustbox}
\usepackage[hidelinks]{hyperref}
\usepackage{todonotes}
\usepackage{makecell}
\usepackage{orcidlink}
\newcommand{\repeatthanks}{\textsuperscript{\thefootnote}}

\begin{document}
\title{SCIsegV2: A Universal Tool for Segmentation of Intramedullary Lesions in Spinal Cord Injury}
\titlerunning{SCIsegV2: Universal Tool for Spinal Cord Injury Lesion Segmentation}
% If the paper title is too long for the running head, you can set an abbreviated paper title here

\author{
Enamundram Naga Karthik\thanks{these authors contributed equally to this work}\inst{1,2}\orcidlink{0000-0003-2940-5514} \and
Jan Valošek\repeatthanks\inst{1,2,3}\orcidlink{0000-0002-7398-4990} \and
Lynn Farner\inst{4}\orcidlink{0009-0004-8252-5448} \and
Dario Pfyffer\inst{4,5}\orcidlink{0000-0002-2406-9251} \and
Simon Schading-Sassenhausen\inst{4}\orcidlink{0000-0003-3969-193X} \and
Anna Lebret\inst{4}\orcidlink{0009-0004-1908-4328} \and
Gergely David\inst{4}\orcidlink{0000-0002-9379-5193} \and
Andrew C. Smith\inst{6}\orcidlink{0000-0001-5020-8094} \and
Kenneth A. Weber II\inst{5}\orcidlink{0000-0002-0916-9174} \and
Maryam Seif\inst{4,7}\orcidlink{0000-0002-9253-5680} \and
RHSCIR Network Imaging Group \and 
Patrick Freund \thanks{joint senior authors} \inst{4,7}\orcidlink{0000-0002-4851-2246} \and
Julien Cohen-Adad \repeatthanks \inst{1,2,8,9}\orcidlink{0000-0003-3662-9532}
}

\institute{
Polytechnique Montreal, Montreal, QC, Canada \and
Mila - Quebec AI Institute, Montreal, QC, Canada \and
Palacký University Olomouc, Olomouc, Czechia \and
University of Zürich, Zürich, Switzerland \and
Stanford University School of Medicine, Stanford, California, USA \and
University of Colorado School of Medicine, Aurora, Colorado, USA \and
Max Planck Institute for Human Cognitive and Brain Sciences, Leipzig, Germany \and
CHU Sainte-Justine, Université de Montréal, Montreal, QC, Canada \and
Functional Neuroimaging Unit, CRIUGM, Université de Montréal, QC, Canada
\email{\{naga-karthik.enamundram,jan.valosek\}@polymtl.ca}
}

% \author{Anonymous}
% \institute{Anonymous Organization}

\authorrunning{Naga Karthik and Valosek et al.}
% \authorrunning{Anonymous et al.}
% First names are abbreviated in the running head.
% If there are more than two authors, 'et al.' is used.

\maketitle              % typeset the header of the contribution
\vspace{-10pt}
\begin{abstract}
Spinal cord injury (SCI) is a devastating incidence leading to permanent paralysis and loss of sensory-motor functions potentially resulting in the formation of lesions within the spinal cord. 
% Structural lesion characteristics acquired at the lesion epicenter based on conventional magnetic resonance imaging (MRI) scans can predict the functional recovery of SCI patients.
%Spinal cord injury (SCI) can lead to permanent physical deficits and significant socio-economic consequences for patients and their families. 
Imaging biomarkers obtained from magnetic resonance imaging (MRI) scans can predict the functional recovery of individuals with SCI and help choose the optimal treatment strategy. 
Currently, most studies employ manual quantification of these MRI-derived biomarkers, which is a subjective and tedious task. 
In this work, we propose (i) a universal tool for the automatic segmentation of intramedullary SCI lesions, dubbed \texttt{SCIsegV2}, and (ii) a method to automatically compute the width of the tissue bridges from the segmented lesion. Tissue bridges represent the spared spinal tissue adjacent to the lesion, which is associated with functional recovery in SCI patients. The tool was trained and validated on a heterogeneous dataset from 7 sites comprising patients from different SCI phases (acute, sub-acute, and chronic) and etiologies (traumatic SCI, ischemic SCI, and degenerative cervical myelopathy).
% The ablation study showed that concatenating the spinal cord segmentation as a second channel, improved performance of the lesion segmentation. On the other hand, aggressive data augmentation  only results in marginal improvements in lesion segmentation performance. \texttt{SCIsegV2} trained on heterogeneous SCI data outperformed phenotype-specific models, which is a crucial step in a suitable long-term solution for segmentation any kind of SCI lesions. 
Tissue bridges quantified automatically did not significantly differ from those computed manually, suggesting that the proposed automatic tool can be used to derive relevant MRI biomarkers. 
\texttt{SCIsegV2} and the automatic tissue bridges computation are open-source and available in Spinal Cord Toolbox (v6.4 and above) via the \texttt{sct\_deepseg -task seg\_sc\_lesion\_t2w\_sci} and \texttt{sct\_analyze\_lesion} functions, respectively.
% \footnote{\href{https://anonymous.4open.science/r/scisegv2-0830}{https://anonymous.4open.science/r/scisegv2-0830}} and will be integrated into the Spinal Cord Toolbox. 
% via \texttt{sct\_deepseg} and \texttt{sct\_analyze\_lesion} functions.

\end{abstract}
%

%\todo{add PRAXIS affiliations}

\section{Introduction}

% \subsection{Intramedullary lesions}

Traumatic and non-traumatic spinal cord injuries (SCI) represent damage to the spinal cord (SC) with severe consequences, including weakness and paralysis in patients \cite{David2019-sw}. Traumatic SCI arises from sudden physical impacts, such as car accidents or falls, while non-traumatic SCI can be caused by ischemia (ischemic SCI) or chronic mechanical compression of the SC (degenerative cervical myelopathy, DCM) \cite{David2019-sw,Ahuja2017-jn}. Both traumatic and non-traumatic SCI commonly involve intramedullary lesions, which are critical areas of tissue damage within the SC. Magnetic resonance imaging (MRI) is routinely used to provide information on the extent and the location of these intramedullary lesions \cite{David2019-sw,Freund2019-sc,Seif2020-ft}. Importantly, MRI scans can also be used to compute quantitative biomarkers, such as midsagittal tissue bridges \cite{Huber2017-ew}. These help in quantifying the amount of preserved SC neural tissue (carrying motor and sensory information to and from the brain) and have been found to predict functional recovery in patients with traumatic and non-traumatic SCI \cite{Huber2017-ew,ODell2020-fo,Pfyffer2019-ah,Pfyffer2021-kf,Smith2022-vp,PFYFFER2024}. 
%Notably, wider tissue bridges measured one month after SCI predicted better recovery at one-year follow-up \cite{Pfyffer2019-ah} and correlated with future walking ability \cite{ODell2020-fo}. 

\vspace{-10pt}
\begin{figure}
\centering
\includegraphics[width=0.95\textwidth]{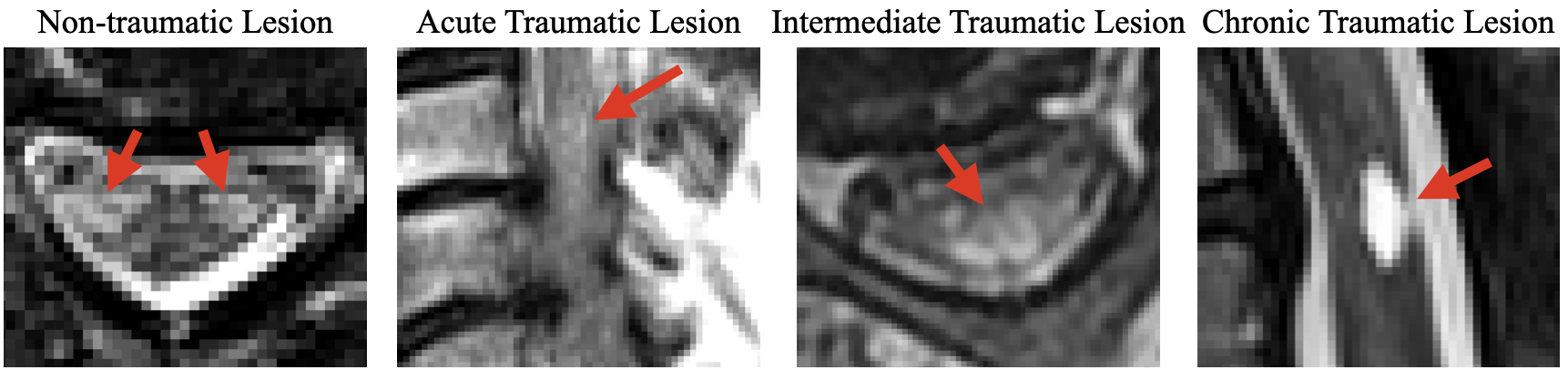}
\caption{Representative axial and sagittal T2w MRI scans of the lesion in various SCI etiologies/types.}
% with their min-max normalized histograms.} 
\label{fig1}
\end{figure}
\vspace{-8pt}

Identifying MRI biomarkers in SCI automatically is challenging due to the varying size and location of lesions across patients and injury phases. Lesions can evolve from hyper- to hypo-intense depending on the underlying pathological mechanism (Figure \ref{fig1}), and metal implants in the spine often cause image artifacts \cite{Freund2019-sc}. As a result, most studies report biomarkers quantified from manual lesion annotations \cite{Huber2017-ew,Vallotton2019-zn,Pfyffer2019-ah,Pfyffer2020-xe,Pfyffer2021-kf,ODell2020-fo,Smith2022-vp,PFYFFER2024}, which is a tedious and error-prone task subject to inter-rater variability, making large-scale studies impractical. While two existing studies proposed automatic lesion segmentation in SCI, their methods were developed for specific SCI etiologies: \cite{McCoy2019-dv} focused on \textit{acute} preoperative traumatic SCI lesions, whereas, \cite{NagaKarthik2024.01.03.24300794} introduced \texttt{SCIseg}, an open-source model for predominantly chronic, post-operative traumatic and ischemic SCI lesions.
% Another important aspect is the \textit{accessibility} of these methods. Open-source models like \cite{NagaKarthik2024.01.03.24300794} enable wider adoption and faster progress in SCI image analysis, leading to unbiased studies on large patient cohorts. 
However, maintaining multiple etiology-specific models is challenging, highlighting the need for a single, \textbf{comprehensive} model for segmenting any kind of SCI lesions.
% Regarding the biomarkers computation, Spinal Cord Toolbox (SCT), an open-source library for spinal cord MRI data analysis [REF SCT], provides functions for automatic lesion volume and lesion length calculation from an input lesion segmentation. Note that the lesion segmentation can be manual or automatic. When combined with SCIseg, SCT provides a convenient way to automatically segment the lesions and quantify their properties (i.e., volume and length). However, there is currently no tool available for the automatic quantification of tissue bridges.  
In this regard, our contributions in this work are as follows: 
% \vspace{-10.6pt}
\begin{itemize}
    \item \textbf{Segmentation tool}: We present a comprehensive tool for the automatic segmentation of intramedullary SCI lesions, dubbed \texttt{SCIsegV2}. Our model was trained and validated on a heterogeneous, multinational dataset from 7 sites consisting of (i) traumatic SCI (acute preoperative, sub-acute and chronic postoperative) and (ii) non-traumatic SCI (ischemic SCI and DCM).
    \item \textbf{Analysis tool}: We propose a method to automatically compute the midsagittal tissue bridges, a predictor of functional recovery in SCI patients. 
    \item \textbf{Packaging in open-source software suite}: Both \texttt{SCIsegV2} and the automatic tissue bridges computation are open-source and will be integrated into the Spinal Cord Toolbox (SCT) v6.4 and higher.
\end{itemize}

\section{Materials and Methods}

\subsection{Dataset}

We used T2-weighted (T2w) MRI images with heterogeneous image resolutions (isotropic, sagittal, and axial) and magnetic field strengths (1.0T, 1.5T, and 3.0T) from seven sites. The number of patients from each site is as follows: 
\{site 1: $(n=154)$, site 2: $(n=80)$, site 3: $(n=14)$, site 4: $(n=11)$, site 5: $(n=23)$, site 6: $(n=4)$, site 7: $(n=5)$\}. Site 1 contained patients with both traumatic $(n=97)$ and non-traumatic SCI (mainly, DCM) $(n=57)$. Sites 2 \& 3 included both preoperative and postoperative traumatic SCI, while sites 4 to 7 included only acute preoperative traumatic SCI.  The timing of the MRI examination in relation to injury for traumatic SCI patients from sites 1, 2, and 3 is detailed in Table 1 of \cite{NagaKarthik2024.01.03.24300794}. 
% patients with traumatic and 57 patients with nontraumatic SCI (DCM); site 2: 80 patients with traumatic SCI; site 3: 14 patients with traumatic SCI; site 4: 11 patients; site 5: 23 patients; site 6: 4 patients, site 7: 5 patients. Sites 1 to 3 included both preoperative and postoperative traumatic SCI, while sites 4 to 7 included only acute preoperative traumatic SCI. 
Eight patients from site 1 were followed up with additional MRI examinations, and 40 patients from site 1 had both sagittal and axial T2w images. 
Patients from sites 1 \& 2 were split according to 80-20\% train/test ratio. Due to the relatively small size of some datasets, we decided to use sites 3, 5 \& 6 entirely for training and kept the patients from sites 4 \& 7 as held-out (unseen) test sets to evaluate the model's generalization performance. 
The model was trained on a total of 281  T2w images and tested on 75 images. 
% (containing both internal and external test sets). 
% We treated all images as independent inputs for the model's training. 
The ground truth masks of intramedullary lesions appearing as T2w signal abnormalities were manually annotated by expert raters at individual sites \cite{NagaKarthik2024.01.03.24300794}. The SC masks were automatically segmented using the \texttt{sct\_deepseg\_sc} \cite{gros2019-deepseg} algorithm and manually corrected when necessary.

\subsection{SCIsegV2}
\label{sec:scisegv2}

We used nnUNet \cite{isensee2021} as the backbone architecture for the SCI lesion segmentation model. 
% The field of medical image segmentation is marked with countless novel architectures, each claiming best performance on specific application domains. While this points to an innovation bias towards newer architectures,
The continuing dominance of nnUNet \cite{isensee2021,isensee2024nnunet} across several open-source challenges has shown that a well-tuned convolutional neural network (CNN) architecture is robust and continues to achieve state-of-the-art results over novel (and more sophisticated) transformer-based architectures in dense pixel prediction tasks such as image segmentation. 
% With recent models building upon the nnUNet framework \cite{Roy2023MedNeXtTS,Ma2024UMambaEL}, there is a trend in employing nnUNet as the standard baseline and proposing improvements upon it.

Similar to the recent work building upon the nnUNet framework \cite{Roy2023MedNeXtTS,Ma2024UMambaEL}, we experimented with its easy-to-tweak trainers for developing \texttt{SCIsegV2}. Specifically, we used: (i) \texttt{nnUNetTrainer}, the default model, and (ii) \texttt{nnUNetTrainerDA5}, the model applying aggressive data augmentation. 
% and (iii) \texttt{nnUNetTrainerDiceCELoss\_noSmooth}, the model that disables the smoothing term in the loss function to stabilize training. 
The augmentation methods in the standard \texttt{nnUNetTrainer} include random rotation, scaling, mirroring, Gaussian noise addition, Gaussian blurring, adjusting image brightness and contrast, low-resolution simulation, and Gamma transformation. 
In addition to the standard augmentations, \texttt{nnUNetTrainerDA5} applies additional transforms such as random patch replacement with mean values, sharpening, median filter, gamma correction, and additive intensity gradients. We note that even though the training data is a collection of heterogeneous datasets (acute, chronic traumatic and non-traumatic SCI) from 7 sites, we employed stronger data augmentation to improve generalization across various types of lesions. 
% The default \texttt{nnUNetTrainer} uses Dice-Cross Entropy loss with the smoothing term set of $1e^{-5}$. With this configuration, we experienced frequent training collapses, especially when the training samples are less and do not contain large lesions (e.g. non-traumatic DCM lesions). To prevent this phenomenon, we used the \texttt{nnUNetTrainerDiceCELoss\_noSmooth} which has a stabilizing effect by setting the smoothing term to zero, thus providing effective gradients through the loss function early in the training.

The wide spectrum of lesion intensities in different SCI etiologies makes lesion segmentation an extremely challenging task. For instance, lesions in acute SCI are mildly T2w hyperintense and chronic lesions are bright T2w hyperintense (Figure \ref{fig1}), although typically surrounded by metallic implants causing heavy interference \cite{Freund2019-sc}. Therefore, we compared two different training strategies for lesion segmentation: (i) given just the T2w image as the input, the model is trained to segment both the SC and lesions in hierarchical ordering, and (ii) assuming the availability of SC segmentation mask, the model is trained to segment only the lesions, given a 2-channel input consisting of the T2w image concatenated with the SC mask. While the first strategy attempts to implicitly guide the model towards the SC for lesion segmentation, the latter provides an explicit guidance in the form of the input channel.

% Lastly, while this is a relatively less known phenomenon, nnUNet experiences frequent crashes when training specifically for lesion segmentation. This is primarily due to the smoothing term in the Dice cross-entropy loss function which is suited towards segmenting larger objects, hence providing better larger gradients to the network during the training phase. As the lesions are typically quite smaller objects, having effect learning signals early in the training is important. Disabling the smoothing term in the loss function, helps in achieving that and hence is crucial. We note that the smoothing term was also disabled in \cite{NagaKarthik2024.01.03.24300794} to obtain a robust traumatic SCI lesion segmentation model. 

\subsection{Automatic Quantification of Tissue Bridges}

\begin{figure}
\centering
\includegraphics[width=0.95\textwidth]{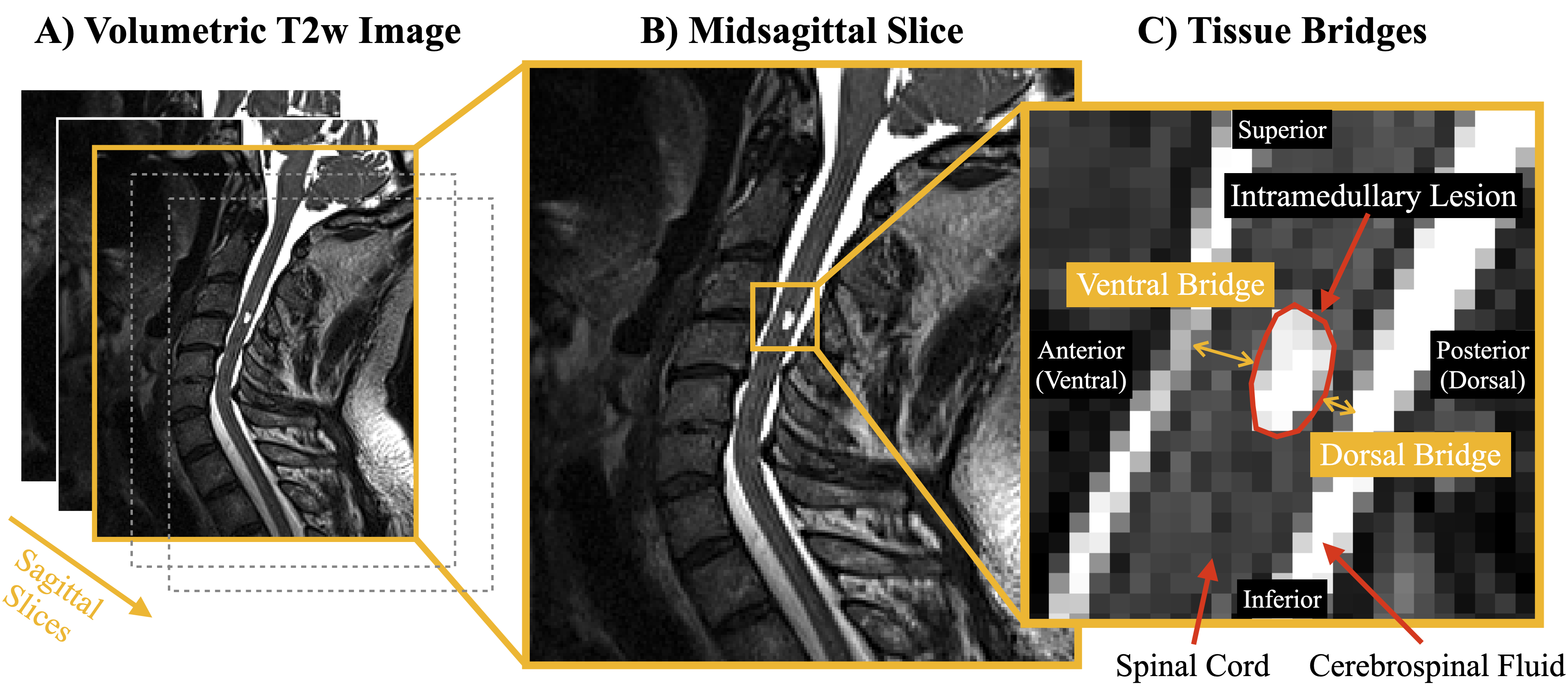}
\caption{Illustration of tissue bridges. A) Volumetric T2w image of a spinal cord injury (SCI) with chronic intramedullary lesion. B) Midsagittal slice used to compute the tissue bridges. C) Ventral and dorsal tissue bridges are defined as the width of spared tissue at the minimum distance from the intramedullary lesion edge to the boundary between the SC and cerebrospinal fluid.} 
\label{fig2}
\end{figure}

The manual measurement of tissue bridges is performed on a single midsagittal slice of a volumetric (3D) T2w MRI image \cite{Huber2017-ew,Vallotton2019-zn,Pfyffer2019-ah,Pfyffer2020-xe,Pfyffer2021-kf,ODell2020-fo,Smith2022-vp,PFYFFER2024} (Figure \ref{fig2}A). The midsagittal slice is defined as the middle slice of all slices where the SC is visible (Figure \ref{fig2}B). Ventral and dorsal tissue bridges are quantified as the width of spared tissue at the minimum distance from the intramedullary lesion edge to the boundary between the SC and cerebrospinal fluid (Figure \ref{fig2}C).

% \vspace{-10pt}
% \vspace{-8pt}

To automate the measurement of tissue bridges, we propose a method that computes ventral and dorsal tissue bridges utilizing the lesion and SC segmentation masks. To compensate for different neck positions and, consequently, different SC curvatures, we use angle correction, which adjusts the tissue bridge widths with respect to the SC centerline \cite{Gros2018-centerline}. 
The method computes tissue bridges from all sagittal slices containing the lesion, allowing quantification of not only midsagittal but parasagittal tissue bridges as well. For the purpose of this study (and to compare against existing manual measurements based on midsagittal tissue bridges), we considered only the midsagittal slice for the automatic measurement of the tissue bridges. 
% The method will be available in the \texttt{sct\_analyze\_lesion} function as part of SCT (v6.4 and above) \cite{de2017sct}.

% \vspace{-5pt}
\subsection{Experiments}
\label{sec:exps}

We divided our experiments into 3 categories to investigate the effects of input types, data augmentation strategies and SCI etiology-specific models. 
% The final experiment focused on comparing manually versus automatically calculated tissue bridges. 
All images were preprocessed with right-left, posterior-anterior, inferior-superior (RPI) orientation, resampled to a common resolution ($0.92 \times 0.68 \times 0.92$ $\textrm{mm}^3$, which is the median of all image resolutions in the training set) and intensity-normalized using Z-score normalization. The model was trained for 1000 epochs with 5-fold cross-validation, using a batch size of 2 and the stochastic gradient descent optimizer with a polynomial learning rate scheduler. 
% All models were trained either with \texttt{nnUNetTrainer} or \texttt{nnUNetTrainerDA5} variants unless otherwise specified.

\vspace{-9pt}
\subsubsection{Inputs} We trained 2 different models: (i) a model that segments \textit{both} the SC and lesions given just the T2w image as input (referred to as \texttt{single}), and (ii) a model that segments \textit{only} the lesion given a 2-channel input consisting of T2w image and the SC segmentation (referred to as \texttt{multi}). As briefly discussed in Section \ref{sec:scisegv2}, this experiment is to understand whether providing additional (localization) context in the form of SC segmentation as input would improve the lesion segmentation performance.

\vspace{-9pt}
\subsubsection{Data augmentation} Given the increasing literature towards unrealistic transformations leading to better test-time performance \cite{billot_synthseg_2023}, we compared 2 models with and without aggressive data augmentation to understand which model leads to better generalization on external test sets with acute preoperative SCI images. These are referred to as \texttt{defaultDA} and \texttt{aggressiveDA}, respectively.

\vspace{-9pt}
\subsubsection{SCI Etiology-specific models} This crucial experiment will provide insight into the initial hypothesis, asking whether a comprehensive SCI model is achievable. Toward this end, we compared our \texttt{SCIsegV2} model trained on all 7 sites against etiology-specific models individually trained on non-traumatic SCI and acute preoperative SCI data, respectively. 
% We note that the non-traumatic SCI model was trained with \texttt{nnUNetTrainerDiceCELoss\_noSmooth} in order to have comparable results.
% This ablation would help understand whether a universal model for SCI lesions is feasible

\vspace{-9pt}
\subsubsection{SCIseg} We also evaluate our model against SCIseg \cite{NagaKarthik2024.01.03.24300794} which was trained on data from three sites comprising traumatic and ischemic SCI lesions. The model used a three-phase training strategy involving active learning and is available in SCT. Details about the training strategy can be found in \cite{NagaKarthik2024.01.03.24300794}.

\vspace{-9pt}
\subsubsection{Tissue Bridges}\label{sec:bridges} To validate the automatic measurements of the tissue bridges, we compared the method against manual and semi-automatic techniques in 15 individuals with traumatic SCI from site 1. Specifically, we compared the following: (1) \textbf{manual} — manual measurement of tissue bridges on manually segmented intramedullary lesions, (2) \textbf{semi-automatic} — automatic measurement of the tissue bridges using the proposed method on manually segmented intramedullary lesions, and (3) \textbf{fully-automatic} — automatic measurement of tissue bridges using \texttt{SCIsegV2} predictions. Statistical analysis was performed using the SciPy v1.10.0. The distribution of the data was assessed with the D’Agostino and Pearson normality test. Subsequently, the Kruskal-Wallis H-test was performed to compare the methods independently for ventral and dorsal bridges.

\vspace{-9pt}
\subsubsection{Evaluation Metrics} We used \texttt{MetricsReloaded} \cite{maier2024metrics} and contributed to its development by adding lesion-wise metrics, specifically, lesion-wise sensitivity, positive predictive value, and $F_1$-score in addition to the existing metrics. 

% Lastly, we evaluated the model's false positive rate by applying the model to an open-source database of healthy controls with \textit{no} intramedullary lesions \cite{Cohen-Adad2021-pn}. 

\section{Results}

\begin{figure}[t!]
\centering
\includegraphics[width=1.025\textwidth]{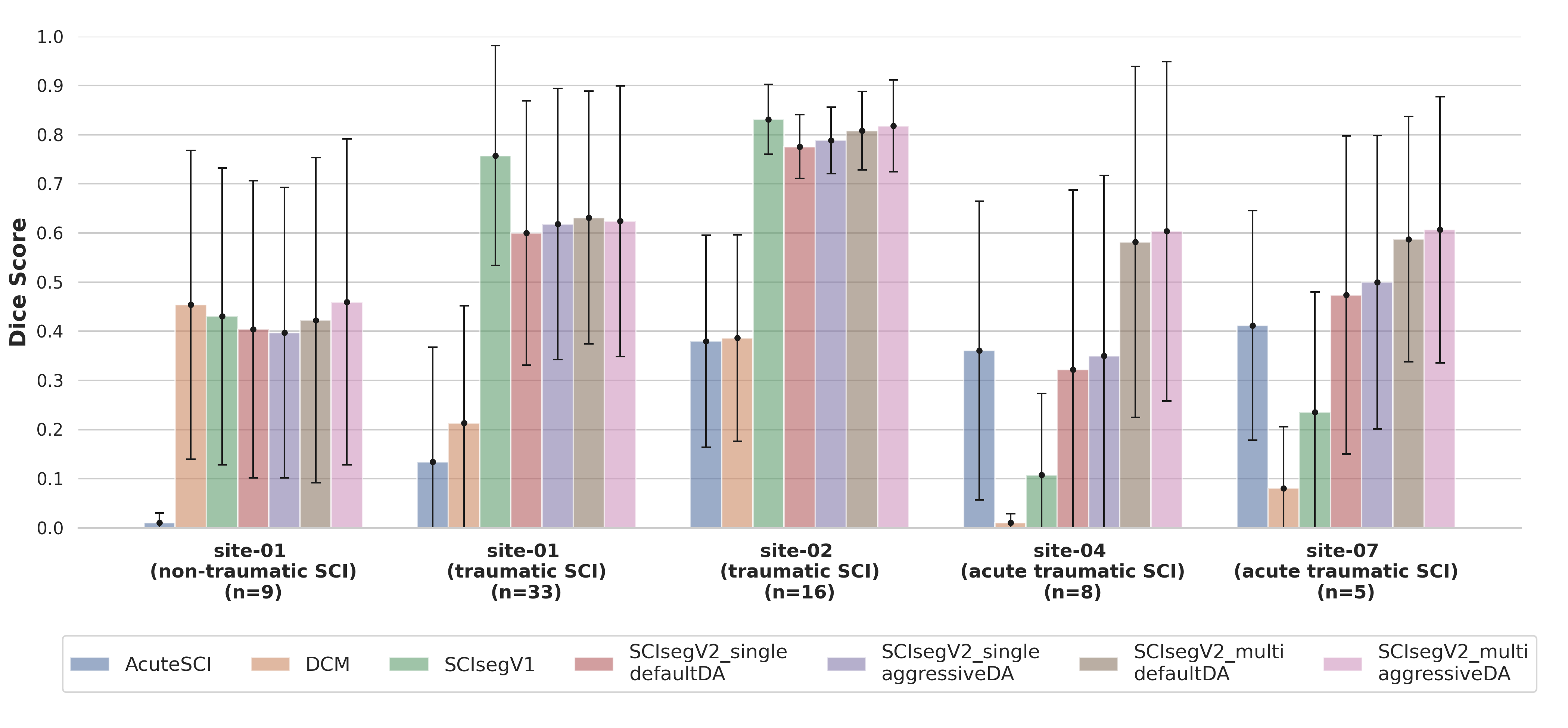}
\caption{Comparison of Dice scores for different SCI models. Each bar plot shows the test Dice scores averaged across 5 folds (the error bar represents the standard deviation).} 
\label{fig3}
\end{figure}

Figure \ref{fig3} shows the test Dice scores averaged across 5 folds for all models described in Section \ref{sec:exps}. Starting with the etiology-specific models, we observed that the model trained only on acute preoperative (\texttt{AcuteSCI}) data does not perform well on sites 1 and 2 containing non-traumatic and traumatic SCI data. While it performs relatively better on test sets from unseen sites (sites 4 and 7), it does not outperform the \texttt{SCIsegV2} models.
Likewise, the model trained only on non-traumatic SCI data (\texttt{DCM}) performs well on a similar test set but fails in generalizing to acute preoperative SCI of sites 4 and 7. 

% \vspace{-10pt}
% \vspace{-10pt}

Since \texttt{SCIsegV1} \cite{NagaKarthik2024.01.03.24300794} was trained on a dataset of 3 sites predominantly consisting of traumatic SCI, it outperforms \texttt{SCIsegV2} models in sites 1 and 2. However, we observed that even \texttt{SCIsegV1} performs poorly on sites 4 and 7 suggesting that the segmentation of acute preoperative SCI lesions is extremely challenging. 
Within \texttt{SCIsegV2} models, we noted that training with aggressive data augmentation  only results in marginal improvements in lesion segmentation performance. However, concatenating the SC segmentation as a second channel along with the input image (resulting in a 2-channel input) showed considerable improvements, especially in acute preoperative images, compared to a single-channel input with just the T2w image. 
% In other words, training the model to segment the lesions by providing explicit guidance in the form the SC segmentation was better than training the model to segment \textit{both} SC and lesion given just the T2w scan as input. A limitation of this approach is the hard requirement of SC segmentation before training the model.

In Table \ref{tab:metrics}, we present a quantitative comparison of different models using lesion-wise metrics. We noticed that the models' performance depends heavily on the specific SCI phases and etiologies. Except for site 1 (traumatic SCI), \texttt{SCIsegV2} outperforms \texttt{SCIsegV1} in all other sites. 
Comparing within the \texttt{SCIsegV2} models, one of downsides of the \texttt{multi}- models, despite achieving higher Dice scores (Figure \ref{fig3}), is its dependency on the SC masks as input. 
% This prevents a direct utilization \texttt{SCIsegV2} for automatically computing the tissue bridges which requires both SC and lesion segmentations.

% \vspace{-10pt}
 \begin{table}[htbp!]
    \centering
    \setlength{\tabcolsep}{4pt} % Adjust the length as needed
    \caption{Comparison of lesion-wise metrics for \texttt{SCIsegV1} and various \texttt{SCIsegV2} models (with \texttt{aggressiveDA}) across the 5 testing sites. Metrics correspond to lesion-wise positive predictive value (PPVL), sensitivity (SensL) and $F_1$-score; higher the value the better ($\uparrow$). For a given site, bold values in each column represent the best model.}
    \resizebox{\textwidth}{!}{%
    \begin{tabular}{llccccc}
    \toprule
        \multirow{3}{*}{\textbf{Model}} & \multirow{3}{*}{\textbf{Metric}} & \multicolumn{5}{c}{\textbf{Test Sites}} \\
        \cline{3-7} & & 
        \multirow{2}{*}{\textbf{\makecell{site-01 \\ (DCM)}}} & \multirow{2}{*}{\textbf{\makecell{site-01 \\ (tSCI)}}} & \multirow{2}{*}{\textbf{\makecell{site-02 \\ (tSCI)}}} & \multirow{2}{*}{\textbf{\makecell{site-04 \\ (acuteSCI)}}} & \multirow{2}{*}{\textbf{\makecell{site-07 \\ (acuteSCI)}}} \\
        \\
        \hline
\multirow{3}{*}{\makecell{\texttt{SCIsegV1}}} 
& $(\uparrow)$  PPVL & \textbf{0.63 $\pm$ 0.43} & \textbf{0.81 $\pm$ 0.31} & 0.95 $\pm$ 0.14 & 0.37 $\pm$ 0.50 & 0.62 $\pm$ 0.50 \\
 & $(\uparrow)$  SensL & 0.78 $\pm$ 0.44 & \textbf{0.95 $\pm$ 0.16} & 0.97 $\pm$ 0.12 & 0.55 $\pm$ 0.49 & 0.72 $\pm$ 0.49 \\
 & $(\uparrow)$  $F_1$ScoreL & 0.67 $\pm$ 0.42 & \textbf{0.84 $\pm$ 0.26} & 0.95 $\pm$ 0.12 & 0.35 $\pm$ 0.47 & 0.64 $\pm$ 0.49 \\
\hline
\multirow{3}{*}{\makecell{\texttt{SCIsegV2} \\ \texttt{single}}} 
& $(\uparrow)$  PPVL & 0.55 $\pm$ 0.45 & 0.73 $\pm$ 0.33 & \textbf{0.96 $\pm$ 0.13} & 0.55 $\pm$ 0.50 & \textbf{0.78 $\pm$ 0.36} \\
 & $(\uparrow)$  SensL & 0.68 $\pm$ 0.49 & 0.91 $\pm$ 0.24 & \textbf{0.97 $\pm$ 0.12} & 0.63 $\pm$ 0.48 & \textbf{0.84 $\pm$ 0.36} \\
 & $(\uparrow)$  $F_1$ScoreL & 0.59 $\pm$ 0.45 & 0.77 $\pm$ 0.28 & \textbf{0.95 $\pm$ 0.12} & 0.53 $\pm$ 0.49 & \textbf{0.80 $\pm$ 0.35} \\
% \hline
% \multirow{3}{*}{\makecell{SCIsegV2 \\ single \\ defaultDA}} 
% & $(\uparrow)$  PPVL & 0.54 $\pm$ 0.44 & 0.73 $\pm$ 0.33 & \textbf{0.98 $\pm$ 0.08} & 0.50 $\pm$ 0.49 & 0.74 $\pm$ 0.40 \\
%  & $(\uparrow)$  SensL & 0.71 $\pm$ 0.48 & 0.89 $\pm$ 0.25 & \textbf{0.97 $\pm$ 0.12} & 0.65 $\pm$ 0.48 & 0.88 $\pm$ 0.27 \\
%  & $(\uparrow)$ F1ScoreL & 0.60 $\pm$ 0.43 & 0.76 $\pm$ 0.28 & \textbf{0.96 $\pm$ 0.11} & 0.50 $\pm$ 0.48 & 0.78 $\pm$ 0.36 \\
\hline
\multirow{3}{*}{\makecell{\texttt{SCIsegV2} \\ \texttt{multi}}} 
& $(\uparrow)$  PPVL & 0.61 $\pm$ 0.41 & 0.70 $\pm$ 0.36 & 0.90 $\pm$ 0.20 & \textbf{0.75 $\pm$ 0.39} & 0.54 $\pm$ 0.41 \\
 & $(\uparrow)$  SensL & \textbf{0.80 $\pm$ 0.42} & 0.90 $\pm$ 0.26 & 0.97 $\pm$ 0.12 & \textbf{0.90 $\pm$ 0.25} & 0.80 $\pm$ 0.42 \\
 & $(\uparrow)$  $F_1$ScoreL & \textbf{0.67 $\pm$ 0.40} & 0.74 $\pm$ 0.32 & 0.91 $\pm$ 0.15 & \textbf{0.75 $\pm$ 0.37} & 0.60 $\pm$ 0.40 \\
% \hline
% \multirow{3}{*}{\makecell{SCIsegV2 \\ multi \\ defaultDA}} 
% & $(\uparrow)$ PPVL & 0.59 $\pm$ 0.40 & 0.73 $\pm$ 0.35 & 0.91 $\pm$ 0.19 & \textbf{0.81 $\pm$ 0.38} & 0.72 $\pm$ 0.34 \\
%  & $(\uparrow)$ SensL & 0.79 $\pm$ 0.42 & 0.90 $\pm$ 0.27 & 0.97 $\pm$ 0.12 & \textbf{0.91 $\pm$ 0.22} & \textbf{0.92 $\pm$ 0.17} \\
%  & $(\uparrow)$ F1ScoreL & 0.65 $\pm$ 0.39 & 0.77 $\pm$ 0.32 & 0.92 $\pm$ 0.15 & \textbf{0.80 $\pm$ 0.36} & 0.78 $\pm$ 0.29 \\
        \bottomrule
        \end{tabular}%
        }
        \label{tab:metrics}
    \end{table}
% \vspace{-10pt}
    
Table \ref{tab:tissue_bridges} shows the comparison of the midsagittal tissue bridges obtained using different methods (manual vs semi-automatic vs fully-automatic; see Section \ref{sec:bridges} for details) for 15 patients with traumatic SCI from site 1. For the fully-automatic technique, we used the \texttt{SCIsegV2\_single\_aggressiveDA} model to obtain the lesion segmentations. There was \textit{no} statistically significant ($p > .05$) difference between the bridges computed using different methods. 

\begin{table}[htbp]
    \centering
    \setlength{\tabcolsep}{5pt} % Adjust the length as needed
    \caption{Comparison of \textit{ventral} and \textit{dorsal} midsagittal tissue bridges between manual, semi-automatic, and automatic measurements. Values are reported in millimetres.}
    \resizebox{\textwidth}{!}{%
        \begin{tabular}{lcccccc}
        \toprule
            \multirow{3}{*}{\textbf{\makecell{ID}}} & 
            \multicolumn{2}{c}{\textbf{\makecell{Manual Lesions \&\\ Manual Measurements}}} & 
            \multicolumn{2}{c}{\textbf{\makecell{Manual Lesions \&\\ Automatic Measurements}}} & 
            \multicolumn{2}{c}{\textbf{\makecell{\texttt{SCIsegV2} Predictions \&\\ Automatic Measurements}}} \\
            % \cline{2-7} &
            \cline{2-3} \cline{4-5} \cline{6-7} &
            {\textbf{\makecell{Ventral }}} & {\textbf{\makecell{Dorsal}}} & 
            {\textbf{\makecell{Ventral}}} & {\textbf{\makecell{Dorsal}}} & 
            {\textbf{\makecell{Ventral}}} & {\textbf{\makecell{Dorsal}}} \\
            \midrule
            sub-zh101 & 0 & 2.65 & 0 & 2.39 & 0.34 & 2.39 \\
            sub-zh102 & 2.10 & 0.83 & 2.25 & 0 & 2.27 & 0.67 \\
            sub-zh104 & 0 & 0 & 0.54 & 0 & 0.55 & 0 \\
            sub-zh105 & 2.70 & 0 & 2.38 & 0.60 & 2.99 & 0 \\
            sub-zh106 & 0 & 0 & 0 & 0 & 0 & 0 \\
            sub-zh107 & 0 & 0.76 & 0 & 0.67 & 0 & 0.65 \\
            sub-zh108 & 1.32 & 0.52 & 1.96 & 0.66 & 2.03 & 0.68 \\
            sub-zh109 & 1.13 & 1.03 & 0.71 & 0 & 1.08 & 0.73 \\
            sub-zh110 & 0 & 0.99 & 0 & 0 & 0 & 0.39 \\
            sub-zh112 & 3.01 & 0.36 & 1.70 & 0.44 & 2.64 & 0.44 \\
            sub-zh114 & 0 & 0 & 0 & 0.38 & 0 & 0 \\
            sub-zh115 & 0 & 0 & 0 & 2.12 & 0 & 0.42 \\
            sub-zh116 & 3.12 & 0.50 & 2.38 & 0 & 2.49 & 0.80 \\
            sub-zh118 & 0.40 & 0 & 0 & 0 & 0 & 0 \\
            sub-zh119 & 2.93 & 2.98 & 1.04 & 0.50 & 1.48 & 0.95 \\
        \bottomrule
        \end{tabular}%
    }
    \label{tab:tissue_bridges}
\end{table}
\vspace{-2pt}

% \vspace{-20pt}
\section{Discussion}

In this work, we proposed \texttt{SCIsegV2}, a DL-based universal tool for the segmentation of intramedullary lesions across different SCI etiologies and phases. We also automated the calculation of midsagittal tissue bridges, a metric representing spared spinal tissue adjacent to the lesion. This metric is relevant as it is associated with functional recovery in individuals with SCI. Both \texttt{SCIsegV2} and the automatic tissue bridges computation are open-source and available in Spinal Cord Toolbox (v6.4 and above) via the \texttt{sct\_deepseg -task seg\_sc\_lesion\_t2w\_sci} and \texttt{sct\_analyze\_lesion} functions, respectively.
% which until now was quantified manually. 
% Compared to previous studies proposing phenotype-specific segmentation methods \cite{McCoy2019-dv,NagaKarthik2024.01.03.24300794}, our findings suggest a good generalization of \texttt{SCIsegV2} on lesions across a broad spectrum of SCI etiologies and phases.

The heterogeneity in the appearance of intramedullary lesions across different SCI phases (acute, sub-acute, chronic) and etiologies (traumatic SCI, ischemic SCI, DCM) makes lesion segmentation extremely challenging, even for trained radiologists. Relatively low prevalence of traumatic SCI and the need for early surgical intervention \cite{Ahuja2017-jn} result in a low number of preoperative MRI scans, which adds to the difficulty in training a robust automatic segmentation model, that performs well on "real world" clinical data across multiple sites. 
%This limits the availability of acute preoperative SCI data, causing our dataset to comprise more postoperative cases.
Moreover, MRI scans of individuals with chronic SCI frequently exhibit image distortions caused by metallic implants, further complicating the segmentation process.

%Therefore, a tool that is capable of segmenting a wide variety of SCI lesions together with the SC allows for a smooth transition from the model's automatic predictions to the computation of tissue bridges, thereby facilitating clinical studies on the impact of various therapeutic strategies for SCI patients. 

% Therefore, we chose to integrate the \texttt{SCIsegV2\_single\_aggressiveDA} segmenting both lesions and SC into SCT.
% No statistically significant difference between manual, semi-automatic, and fully-automatic measurements of tissue bridges could imply that our method can be used with \texttt{SCIsegV2}'s predictions to obtain tissue bridges fully automatically within comparable range to that of the manual measurements.  
% This is clinically relevant as it would save massive amounts of time for clinical researchers in SCI \cite{Farner2024}. 

As a way of simplifying the lesion segmentation problem, using the SC mask to explicitly guide the model towards the cord showed improved results on certain etiologies. However, it introduced a dependency on the SC mask to be concatenated to the input image, preventing a smooth transition from the model's automatic predictions to the computation of tissue bridges, which requires both the cord and lesion masks. 
In contrast, the \texttt{SCIsegV2\_single} model capable of segmenting both SC and lesions has a higher utility as it is also applicable in scenarios where SC masks are unavailable.
Lastly, as the lesion appearance varies substantially across different SCI types, universal models like \texttt{SCIsegV2} learn the differences in lesion distributions across etiologies and even outperform etiology-specific models while showing good generalization across sites. 

\vspace{-13pt}
\subsubsection{Limitations and Future Work} One limitation of the model is its higher sensitivity to traumatic SCI lesions, as approximately ($\sim 70\%$) of our dataset consists of this population. This skew is due to the relatively low availability of acute preoperative SCI scans. Acquiring additional data in SCI is generally challenging, but is possible with the help of ongoing clinical trials and consortiums for creating large databases for SCI research \cite{Noonan2011,Birkhausere039164}. 
While this work used a relatively small and unbalanced cohort, we presented a preliminary proof-of-concept toward a universal tool for SCI lesion segmentation. 
% Future studies could benefit from larger (and more balanced) cohorts. 
The current literature only quantifies tissue bridges manually, from a single \textit{midsagittal} slice. However, as lesions are 3D blob-like objects, the midsagittal slice might not necessarily contain the largest portions of the lesion and does not consider parasagittally running fiber tracts. Therefore, combining both \textit{parasagittal} and \textit{midsagittal} slices could provide a comprehensive evaluation of the width of the spared tissue bridges. 
% which we will consider for future work.

% \begin{itemize}
%     \item soft lesions  (really? :p)
%     \item para-sagittal slices in tissue bridges
%     \item handle more than one lesion in tissue bridges
%     \item more acute SCI images (currently, dataset is skewed towards tSCI)
%     \item less of number training/testing samples in acute SCI -- hard to get large datasets of acute preop images
% \end{itemize}

\vspace{-12pt}
\subsubsection{Prospect of application} 
Automatic segmentation of the lesions and spinal cord could mitigate the bottleneck and inter-rater variability associated with manual annotations. 
Automating the measurements of tissue bridges could provide an objective, unbiased way in guiding rehabilitation decision making and stratifying patients into homogeneous subgroups of recovery in clinical trials.
%, leading to improved accuracy of prognosis after SCI. 
% considering both mid- and parasagittal slices could provide a more precise evaluation of patients' potential recovery and help choose the optimal treatment strategy. 
% help to further optimize identification of SCI patients' subgroups in future studies and clinical trials.
% With the integration of \texttt{SCIsegV2} in SCT, clinical researchers in SCI community have access to the state-of-the-art model. Moreover, the computation of midsagittal tissue bridges, which was so far done manually, can now be quickly computed using \texttt{sct\_analyze\_lesion}, saving the researchers a lot of time and enabling them to obtain unbiased assessments of the tissue bridges. \ldots

\begin{credits}
\subsubsection{\ackname} 
We thank Mathieu Guay-Paquet and Joshua Newton for their assistance with the management of the datasets and the implementation of the algorithm to SCT. We thank Maxime Bouthillier for the help with manual annotations. We thank Dr. Serge Rossignol and the Multidisciplinary Team on Locomotor Rehabilitation (Regenerative Medicine and Nanomedicine, CIHR), and all the patients. 
The authors would also like to thank the RHSCIR participants and network, including all the participating local RHSCIR sites: Vancouver General Hospital, Foothills Hospital, Royal University Hospital, Toronto Western Hospital, St. Michael’s Hospital, Sunnybrook Health Sciences Centre, Hamilton General Hospital, The Ottawa Hospital Civic Campus, Hôpital de l’Enfant Jésus, Hôpital du Sacre Coeur de Montréal, QEII Health Sciences Centre, Saint John Regional Hospital. 

Funded by the Canada Research Chair in Quantitative Magnetic Resonance Imaging [CRC-2020-00179], the Canadian Institute of Health Research [PJT-190258], the Canada Foundation for Innovation [32454, 34824], the Fonds de Recherche du Québec - Santé [322736, 324636], the Natural Sciences and Engineering Research Council of Canada [RGPIN-2019-07244], the Canada First Research Excellence Fund (IVADO and TransMedTech), the Courtois NeuroMod project, the Quebec BioImaging Network [5886, 35450], INSPIRED (Spinal Research, UK; Wings for Life, Austria; Craig H. Neilsen Foundation, USA), Mila - Tech Transfer Funding Program, the Association Française contre les Myopathies (AFM), the Institut pour la Recherche sur la Moelle épinière et l'Encéphale (IRME), the National Institutes of Health Eunice Kennedy Shriver National Institute of Child Health and Development (R03HD094577). ACS is supported by the National Institutes of Health – K01HD106928 and R01NS128478 and the Boettcher Foundation’s Webb-Waring Biomedical Research Program. KAW is supported by the National Institutes of Health – K23NS104211, L30NS108301, R01NS128478. 
The Rick Hansen Spinal Cord Injury Registry and this work are supported by funding from the Praxis Spinal Cord Institute through the Government of Canada and the Province of British Columbia. For more information about RHSCIR(9), please visit \href{www.praxisinstitute.org}{www.praxisinstitute.org}.
JV received funding from the European Union's Horizon Europe research and innovation programme under the Marie Skłodowska-Curie grant agreement No 101107932 and is supported by the Ministry of Health of the Czech Republic, grant nr. NU22-04-00024. ENK is supported by the Fonds de Recherche du Québec Nature and Technologie (FRQNT) Doctoral Training Scholarship. The authors thank Digital Research Alliance of Canada for the compute resources used in this work.

% Site and Participant Acknowledgements

\subsubsection{\discintname}
The authors have no competing interests to declare that are
relevant to the content of this article. 
\end{credits}

\begin{comment}
\end{comment}
%
% ---- Bibliography ----
%
% BibTeX users should specify bibliography style 'splncs04'.
% References will then be sorted and formatted in the correct style.
%
\bibliographystyle{splncs04}
\bibliography{references}

\end{document}